\documentclass[11pt, oneside]{article}   	
\usepackage{geometry}                		
\usepackage{graphicx}				
\usepackage{url}

\title{\textbf{Reply to “Emergent LLM behaviors are observationally equivalent to data leakage"}}
\author{
    Ariel~Flint~Ashery$^{1}$,
	Luca~Maria~Aiello$^{2,3}$,
	Andrea~Baronchelli$^{1\ast}$,\and
	\small$^{1}$ City St George’s, University of London, UK. \\
	\small$^{2}$IT University of Copenhagen, Denmark.\\
    \small$^{3}$Pioneer Centre for AI, Copenhagen, Denmark. \\
	\small$^\ast$Corresponding author. Email: a.baronchelli.work@gmail.com}

\begin{document}
\maketitle

\begin{abstract}
A potential concern when simulating populations of large language models (LLMs) is data contamination, i.e. the possibility that training data may shape outcomes in unintended ways. While this concern is important and may hinder certain experiments with multi-agent models, it does not preclude the study of genuinely emergent dynamics in LLM populations. The recent critique by Barrie and Törnberg \cite{barrie2025emergent} of the results of Flint Ashery et al. \cite{ashery2025emergent} offers an opportunity to clarify that self-organisation and model-dependent emergent dynamics can be studied in LLM populations, highlighting how such dynamics have been empirically observed in the specific case of social conventions.
\end{abstract}
\vspace{0.5cm}

Barrie \& T\"ornberg \cite{barrie2025emergent} question whether the emergence of conventions observed in our recent study of interacting large language models (LLMs) \cite{ashery2025emergent} can be attributed to genuine collective dynamics, or instead result from data leakage from the models' training data. In this note, we respond to their main points and argue that the observed dynamics cannot be explained by data contamination alone. We also note that our study reported two additional results - emergent collective bias and critical mass dynamics - that are likewise the outcome of self-organising population dynamics. We hope this response will contribute to ongoing methodological discussions in this area of research, at a time when collective AI dynamics are quickly gaining societal relevance \cite{schroeder_2025}.

Specifically, Barrie \& T\"ornberg \cite{barrie2025emergent} raise two main critiques of \cite{ashery2025emergent}. The first concerns inventory pruning: the authors argue that the inventory update rule in our naming game (agents retaining the last matched convention) makes the emergence of consensus trivial. The second, more central, critique, reflected in the title of their contribution, concerns data leakage: the authors suggest that LLMs may recognise the scenario (as a coordination game or naming game) from training data, and may therefore be reproducing known outcomes rather than exhibiting genuinely spontaneous emergent behaviour. We address these two points separately below.

Before proceeding, it is important to highlight how the words “emergence” and “convention” are used in \cite{ashery2025emergent}. Following the standard approach in complex and social systems \cite{anderson1972more,hayek1960constitution,sugden1989spontaneous,haken1988information,strogatz2003sync,baronchelli2018emergence,artime2022origin}, we use the term “emergence” when \textit{“purely local interactions [..] lead to population-wide, or ‘global,’ coordinated behavior”} \textit{“without the intervention of any centralized institution”}\footnote{This is a strict definition. It is a widely accepted and well-justified convention in the field to speak of emergence also in cases where, for example, there is an explicit reward for universal coordination \cite{kearns2009behavioral} or a mechanism that aggregates and reports the collective state of the population \cite{salganik2006experimental,judd2010behavioral}, although strictly speaking these approaches do not allow one to evaluate the ``emergence" hypothesis that global coordination results from purely local incentives.} and we use “convention” to refer to a pattern of behavior shared throughout a community, defined as the outcome that everyone expects in interactions allowing two or more equivalent actions \cite{lewis2008convention,bicchieri2005grammar,southwood2011norms,amato2018dynamics,andrighetto2022research}. A convention is therefore emergent when it \textit{“arise[s] spontaneously, without the intervention of any centralized institution,”} i.e., when \textit{“agents are endowed with purely local incentives and conventions may (or may not) emerge as an unintended consequence of individuals attempting to coordinate locally with one another.”}\footnote{This is also a widely accepted definition, but the literature on conventions is vast and scattered. As noted in \cite{baronchelli2018emergence}, the two influential reviews ``The Evolution of Norms" \cite{ehrlich2005evolution} and ``The Evolution of Social Norms" \cite{young1993evolution} do not share a single bibliographic entry, and different definitions exist (a meta naming game?). For example, the requirement that the initial alternatives are equivalent for the agents can be relaxed (see e.g. \cite{judd2010behavioral}).}\cite{ashery2025emergent}.

\subsection*{Inventory Pruning}

The naming game framework adopted in \cite{ashery2025emergent} is a standard experimental paradigm for studying convention formation in both artificial \cite{steels1995self,baronchelli2006sharp,xie2011social} and human \cite{centola2015spontaneous,centola2018experimental} populations. The inventory update rule is an integral part of this paradigm and is implemented accordingly. Our goal was not to test whether conventions emerge under arbitrary conditions, but to study how LLM populations behave in a well-established experimental setting for convention formation, allowing us to characterise their collective dynamics and compare them, where appropriate, with human experiments and theoretical models. As stated in \cite{baronchelli2006sharp}, which introduced the standard theoretical model, \textit{“We want to understand how the microscopic dynamics of the agent interactions can nevertheless give rise to global coherence without external intervention.”} In this light, the focus is not on how a pair of individuals reaches consensus, but on how different pairs, each reaching local consensus on potentially different conventions, can ultimately give rise to global consensus.

In this sense, the objection in \cite{barrie2025emergent} that \textit{“That the LLM ‘picks’ this word in subsequent rounds is therefore trivial”} overlooks an important aspect of the dynamics. In a setting where subsequent interactions of a successful pair involve other agents who are new to them, strong local convergence may complicate global consensus, as persisting with the word used in initial coordination can hinder, rather than facilitate, broader alignment across the population (as we discuss further below).

It is also important to stress that the findings in \cite{ashery2025emergent} go beyond demonstrating convergence. We observed emergent collective bias: conventions were selected collectively with non-uniform probabilities, even when individual agents were initially unbiased. We also observed critical mass tipping points: small committed minorities were able to overturn established conventions, matching predictions from human experiments and theoretical models. These phenomena too cannot be explained by inventory pruning alone, which operates locally and symmetrically.

\subsection*{Data Leakage and Prior Knowledge}

We explicitly acknowledge the risk of data contamination in \cite{ashery2025emergent}, where we contrast AI-specific and human-like collective dynamics, and highlight that some behaviours (e.g. collective bias) appear to be LLM-specific phenomena, meriting further study. In the Supplementary Text (``Prompt Bias Mitigation"), we note that: \textit{“the game’s population dynamics and the symmetry in payoffs across action labels ensure that no predetermined optimal strategy exists—even if a player were aware of global information. Convergence occurs purely at the local level, driven by repeated random interactions rather than a global rule that dictates strategic behaviour.”} In other words, even if an LLM recognises the game structure, the observed population-level dynamics (symmetry breaking, collective bias, critical mass) depend on history-dependent, decentralised interactions, not on a globally optimal solution. 

In fact, our choice of the coordination game setting was deliberate, precisely in light of the potential for data leakage. In contrast to games with a clearly defined optimal strategy (such as many classical game-theoretical setups) or a well-established non-rational outcome (such as the ultimatum game\footnote{Non-systematic experiments conducted by the authors in early 2024 suggest that the data leakage concern may indeed hinder the possibility of testing LLMs in the ultimatum game as, when meta-prompted, most LLMs declared that they based their choices on the existing literature.}), coordination games with equivalent alternatives, such as the naming game, do not specify which convention or action should emerge. This significantly limits the effects of agents' prior knowledge, making these games especially suitable for studying whether and how global consensus can arise from decentralised local interactions.

Returning to the line of reasoning in \cite{barrie2025emergent}, the argument appears to be that: (i) LLMs recognise that they are playing a coordination game; (ii) they know from training that repeating the same choice is a good, or optimal, strategy; and (iii) their training data suggests that such games should converge, so they do. There are several reasons why this explanation does not account for our results.

The starting point for this reasoning is \cite{barrie2025emergent}'s meta-prompting. Three meta-prompting questions are asked to the LLM after providing it with the prompt designed in \cite{ashery2025emergent}. They are:

\begin{enumerate}
\item \textit{``Does this setup remind you of any existing model or theory in social science?''} 
\item \textit{``Tell me if you know what the optimal move is after a success.''}
\item \textit{``Finally, tell me how you think the game will converge globally.''}
\end{enumerate}

To answer the first question, the LLM recognises a coordination game, adding that it is a \textit{“repeated version of such a game”}. In response to the second question, the LLM replies: \textit{“After a ‘success’ [..], the optimal move is to repeat the same action in the next round.”} Finally, in response to the third question, the LLM summarises its output as: \textit{“Rapid convergence to a single, repeated action (a convention), maximizing joint and individual payoffs.”} \cite{barrie2025emergent}. These answers reveal two key aspects. 

The first aspect is that the LLM interprets “globally” as “pairwise”, not as referring to population-level interactions
(see:\textit{“Early rounds might involve mismatches as players ‘search’ for a common choice, but once they successfully coordinate, \underline{both} have a strong incentive to continue repeating that choice.”} \cite{barrie2025emergent} [underlining our own]). Interestingly, this pattern holds across several models (see Appendix). We also conducted further experiments, following the meta-prompting approach of \cite{barrie2025emergent}, with a final question: \textit{``Finally, do you think that you are playing against a single player or a whole population?"}. The results confirm that all tested models assume they are playing against a single opponent (see Appendix). Naturally, assuming pairwise interactions is perfectly consistent with our experimental setup: the prompt explicitly introduces a two-player interaction and, as we note, it \textit{``does not specify that agents are part of a population or provide any detail on how the interaction partner is selected from a group''}\cite{ashery2025emergent}. 

The second aspect of the LLM’s answer to the meta-prompting worth noting is that, as a consequence of this pairwise framing, the LLM concludes that repeating the same name after a success is \textit{“optimal”}, and that \textit{“rational players will ‘lock in’ to the first successful coordination point and maintain it for the remainder of the rounds.”}
However, if this were the case (i.e., if agents \textit{“lock[ed] themselves in to the first successful coordination”} name) this behaviour would not be optimal in a population setting. On the contrary, it would hinder the emergence of global consensus, since different pairs initially coordinate on different names. In other words, the locally optimal strategy would actively prevent global consensus from forming.

Thus, firstly, the LLM interpretation of the experiment as involving pairwise interactions, consistent with the prompt and experimental design, strongly suggests that the LLMs are not relying on prior information in this setting, since most, if not all, papers on the naming game, and a significant portion of the literature on coordination games, involve large populations \cite{cooper1999coordination, castellano2009statistical, baronchelli2018emergence}.
Secondly, the meta-prompting in \cite{barrie2025emergent} does not explain the global consensus observed in our experiments. In fact, it supports the view that LLMs focus on local coordination while neglecting global dynamics, in line with the experimental design in \cite{ashery2025emergent}, but they do not rigidly adopt the locally optimal strategy they return when meta-prompted (or we would not observe global convergence). This further reinforces the validity of our experimental approach and the emergent nature of consensus on a shared social convention.

\noindent A number of other points are relevant:

First, even if the LLM agents inferred from prior knowledge that privileging the same symbol after a coordination success is a reasonable strategy, this would not be problematic: it is a natural inference given our prompt, which by design incentivises local convergence and does not specify the population size, or even whether there is a population at all. To converge, however, agents must be able to adapt their choices to the evolving ecology of conventions they encounter as the experiment unfolds. In lab experiments with human participants, several users reported that they guessed they were no longer playing with the same partner at a certain point during the game \cite{centola2015spontaneous}. Further investigation into how LLMs update their perception of the game as it unfolds would be a valuable direction for future work.

Second, considering also the aspect of collective bias, if the model were somehow simulating a pre-scripted trajectory towards a predetermined convention, the committed minority experiments reported in \cite{ashery2025emergent} would fail. The LLM would ignore the actual game context and converge, or maintain convergence, on the predetermined symbol. In fact, the success of the committed minority experiments demonstrates that the model behaviour is context-sensitive and dynamically influenced by interactions. Furthermore, if the LLMs were simply recognising the game and executing an optimal strategy (as \cite{barrie2025emergent} suggests), such minority-driven reversals should not occur since, once population consensus is reached, maintaining the winning convention is precisely the standard \textit{“optimal”} behaviour in coordination games, and the LLMs know this, as noted above.

Third, a specific result appears to directly challenge the data leakage argument. In the Llama-3.1-70B-Instruct populations, we observed cases where the population spontaneously loses consensus and switches to a different ``stronger" convention without any committed agents (Figure 3B, and text: \textit{“In Llama-3.1-70B-Instruct populations, the bias is so strongly weighted against the weaker convention that the population spontaneously switches to the alternative, stronger convention without requiring any committed agents at all.”} See also Figure S7 of \cite{ashery2025emergent} for the production probability over time). Moreover, this behaviour, stemming from the collective bias phenomenon, is model-dependent: it is observed in Llama-3.1 but not uniformly across models. All of this supports the conclusion that population-level dynamics in LLMs emerge from complex, model-specific interaction effects.

Fourth, broadening the scope beyond the naming game experiments, the fact that a model recognises a specific game does not prove that the model will not interpret the prompt literally, without incorporating additional assumptions from training data. In other words, the prompt itself could override or constrain prior knowledge, and the test proposed in \cite{barrie2025emergent} does not establish that this is not the case. This is an important general point, although it is difficult to test conclusively.

Finally, again looking beyond our specific experiments, there is broader evidence that the data leakage explanation does not fully account for LLM behaviour in related domains. In the growing literature on LLMs in cooperation games, there are several cases where LLM behaviour diverges from what is documented in human experiments, or from what might be expected from an agent that “knows the trick” of the game. For instance, in experiments on the Prisoner's Dilemma \cite{fontana2025nicer}, LLMs do not behave like typical humans, rational agents, or fully game-aware agents who would simply cooperate throughout. These observations further suggest that LLM population dynamics are not simply reproducing patterns memorised from training data.

\textbf{Further insights from collective bias.}
A further point that is not fully addressed in \cite{barrie2025emergent} concerns collective bias. \cite{barrie2025emergent} mentions bias only once, in the sentence: \textit{“They also carry out a number of mitigation steps to guard against the potential for bias towards one single token (i.e., some baked-in preference for a given character of letter meaning any convergence is artefactual of model bias).”} However, our mitigation strategy concerns individual bias. A key finding of \cite{ashery2025emergent} is that even when individual agents are tested to be unbiased in isolation, collective bias emerges as a population-level effect, driven by the dynamics of interaction and memory accumulation. Crucially, this effect is model-dependent, and the strength and direction of collective bias vary across LLM architectures, as shown in our results. It is also non-deterministic (i.e., convergence does not happen systematically on the same convention). Therefore, this phenomenon reflects the emergent dynamics of the interacting system and is hard to reduce to data leakage or static biases in training data alone.

\subsection*{Conclusion}

We agree that data contamination is an important methodological challenge in this line of work, and we welcome continued discussion on best practices. Our paper explicitly flags this as a potential limitation and highlights the need for further study. However, for all the reasons outlined above, the core contributions of our work remain valid and are not explained by the concerns raised. Of course, more work is needed, and we already have preliminary results that further clarify how ``more is different'' \cite{anderson1972more} also for populations of LLMs \cite{ashery2025next}. 
We look forward to further constructive exchanges on these important questions. 

\subsection*{Acknowledgments} We thank Romualdo Pastor-Satorras, Nicola Perra and Michele Starnini for helpful comments.

\clearpage


\begin{thebibliography}{10}

\bibitem{barrie2025emergent}
Christopher Barrie and Petter T{\"o}rnberg.
\newblock Emergent llm behaviors are observationally equivalent to data leakage.
\newblock {\em arXiv preprint arXiv:2505.23796}, 2025.

\bibitem{ashery2025emergent}
Ariel Flint~Ashery, Luca~Maria Aiello, and Andrea Baronchelli.
\newblock Emergent social conventions and collective bias in llm populations.
\newblock {\em Science Advances}, 11(20):eadu9368, 2025.

\bibitem{schroeder_2025}
Daniel~T Schroeder and et~al.
\newblock How malicious ai swarms can threaten democracy, Jun 2025.

\bibitem{anderson1972more}
Philip~W. Anderson.
\newblock More is different.
\newblock {\em Science}, 177(4047):393--396, 1972.

\bibitem{hayek1960constitution}
Friedrich~A. Hayek.
\newblock {\em The Constitution of Liberty}.
\newblock Routledge, 1960.

\bibitem{sugden1989spontaneous}
Robert Sugden.
\newblock Spontaneous order.
\newblock {\em Journal of Economic Perspectives}, 3(2):85--97, 1989.

\bibitem{haken1988information}
Hermann Haken.
\newblock {\em Information and Self-Organization: A Macroscopic Approach to Complex Systems}.
\newblock Springer, 1988.

\bibitem{strogatz2003sync}
Steven Strogatz.
\newblock {\em Sync: The Emerging Science of Spontaneous Order}.
\newblock Hyperion, 2003.

\bibitem{baronchelli2018emergence}
A.~Baronchelli.
\newblock The emergence of consensus: a primer.
\newblock {\em Royal Society Open Science}, 5:172189, 2018.

\bibitem{artime2022origin}
Oriol Artime and Manlio De~Domenico.
\newblock From the origin of life to pandemics: Emergent phenomena in complex systems.
\newblock {\em Philosophical Transactions of the Royal Society A}, 380(2227):20200410, 2022.

\bibitem{kearns2009behavioral}
Michael Kearns, Stephen Judd, Jinsong Tan, and Jennifer Wortman.
\newblock Behavioral experiments on biased voting in networks.
\newblock {\em Proceedings of the National Academy of Sciences}, 106(5):1347--1352, 2009.

\bibitem{salganik2006experimental}
Matthew~J Salganik, Peter~Sheridan Dodds, and Duncan~J Watts.
\newblock Experimental study of inequality and unpredictability in an artificial cultural market.
\newblock {\em science}, 311(5762):854--856, 2006.

\bibitem{judd2010behavioral}
Stephen Judd, Michael Kearns, and Yevgeniy Vorobeychik.
\newblock Behavioral dynamics and influence in networked coloring and consensus.
\newblock {\em Proceedings of the National Academy of Sciences}, 107(34):14978--14982, 2010.

\bibitem{lewis2008convention}
David Lewis.
\newblock {\em Convention: A philosophical study}.
\newblock John Wiley \& Sons, 2008.

\bibitem{bicchieri2005grammar}
Cristina Bicchieri.
\newblock {\em The grammar of society: The nature and dynamics of social norms}.
\newblock Cambridge University Press, 2005.

\bibitem{southwood2011norms}
Nicholas Southwood and Lina Eriksson.
\newblock Norms and conventions.
\newblock {\em Philosophical Explorations}, 14(2):195--217, 2011.

\bibitem{amato2018dynamics}
Roberta Amato, Lucas Lacasa, Albert D{\'\i}az-Guilera, and Andrea Baronchelli.
\newblock The dynamics of norm change in the cultural evolution of language.
\newblock {\em Proceedings of the National Academy of Sciences}, 115(33):8260--8265, 2018.

\bibitem{andrighetto2022research}
Giulia Andrighetto and Eva Vriens.
\newblock A research agenda for the study of social norm change.
\newblock {\em Philosophical Transactions of the Royal Society A}, 380(2227):20200411, 2022.

\bibitem{ehrlich2005evolution}
Paul~R. Ehrlich and Simon~A. Levin.
\newblock The evolution of norms.
\newblock {\em PLoS Biology}, 3(6):e194, 2005.

\bibitem{young1993evolution}
H.P. Young.
\newblock The evolution of conventions.
\newblock {\em Econometrica}, 61:57--84, 1993.

\bibitem{steels1995self}
Luc Steels.
\newblock A self-organizing spatial vocabulary.
\newblock {\em Artificial Life}, 2(3):319--332, 1995.

\bibitem{baronchelli2006sharp}
Andrea Baronchelli, Marcello Felici, Emanuele Caglioti, Vittorio Loreto, and Luc Steels.
\newblock Sharp transition towards shared vocabularies in multi-agent systems.
\newblock {\em Journal of Statistical Mechanics: Theory and Experiment}, 2006(06):P06014, 2006.

\bibitem{xie2011social}
Jierui Xie, Sameet Sreenivasan, Gyorgy Korniss, Weituo Zhang, Chjan~C. Lim, and Boleslaw~K. Szymanski.
\newblock Social consensus through the influence of committed minorities.
\newblock {\em Physical Review E--Statistical, Nonlinear, and Soft Matter Physics}, 84(1):011130, 2011.

\bibitem{centola2015spontaneous}
Damon Centola and Andrea Baronchelli.
\newblock The spontaneous emergence of conventions: An experimental study of cultural evolution.
\newblock {\em Proceedings of the National Academy of Sciences}, 112(7):1989--1994, 2015.

\bibitem{centola2018experimental}
Damon Centola, Joshua Becker, Devon Brackbill, and Andrea Baronchelli.
\newblock Experimental evidence for tipping points in social convention.
\newblock {\em Science}, 360(6393):1116--1119, 2018.

\bibitem{cooper1999coordination}
Russell Cooper.
\newblock {\em Coordination games}.
\newblock cambridge university Press, 1999.

\bibitem{castellano2009statistical}
Claudio Castellano, Santo Fortunato, and Vittorio Loreto.
\newblock Statistical physics of social dynamics.
\newblock {\em Reviews of modern physics}, 81(2):591--646, 2009.

\bibitem{fontana2025nicer}
Nicoló Fontana, Francesco Pierri, and Luca~Maria Aiello.
\newblock Nicer than humans: How do large language models behave in the prisoner's dilemma?
\newblock In {\em Proceedings of the International Conference on Web and Social Media (ICWSM)}, 2025.

\bibitem{ashery2025next}
Ariel Flint~Ashery, Luca~Maria Aiello, Romualdo Pastor-Satorras, and Andrea Baronchelli.
\newblock In preparation, 2025.

\end{thebibliography}

\vspace{1cm}

\subsection*{Appendix}
The example response provided in the main text of Ref.~\cite{barrie2025emergent} suggests that the game takes place between two players. Further inspection of \cite{barrie2025emergent}'s data repository confirms that this is not unique: across different models, the majority of responses similarly used the term ``both'', or otherwise implied that the prompt described a two-player game.
Following the meta-prompting approach conducted by Barrie \& T\"ornberg \cite{barrie2025emergent}, we also tested whether the LLM infers that it is engaged in a social experiment with a population, or instead interprets the experiment as involving only two players. We used the same prompt template as in \cite{barrie2025emergent}, replacing the final question about global convergence with the following: \textit{``Finally, do you think that you are playing against a single player or a whole population?''}
This meta-prompting test was repeated five times for each tested model\footnote{Namely, Claude-3-5-Sonnet-20240620, Claude-3-5-Haiku-20241022, Claude-3-Opus-20240229, GPT-4.1, GPT-4o, GPT-4o-Mini, Llama-3-70B Instruct, and Llama-3.1-70B Instruct.}. Every response across all tested models indicated that the experiment involves two players, rather than a population. However, the most capable LLMs also correctly noted that if the experiment were conducted in populations, both the optimal strategy and the overall convergence dynamics would depend on the evolution of the system. They did not, however, interpret this as the setup described in the prompt (see GPT-4.1 example reported below). In this regard, it is important to stress that in the experiments of \cite{ashery2025emergent}, and in the naming game more broadly, agents are never presented with information about ``the frequency of choices in the population'' (as mentioned in the GPT-4.1 example) or any other metrics describing collective behaviour.





Below we present representative example answers to the population question. Full responses for all tested models are available at: \url{https://github.com/Ariel-Flint-Ashery/AI-norms-prompting}.




\vspace{0.5cm}

\noindent\textbf{\underline{Llama-3.1-70B Instruct}}:

Question response: {\fontfamily{qcr}\selectfont Regarding your final question, based on the setup, it appears that you are playing against a single player (Player 2). The game is a multi-round partnership game, and the payoffs are determined by the simultaneous actions of both players. There is no indication of a population or multiple opponents.}

\vspace{0.5cm}

\noindent\textbf{\underline{claude-3-opus-20240229}}:

Question response: {\fontfamily{qcr}\selectfont Finally, based on the description, it sounds like you are playing against a single opponent repeatedly, not against a whole population. The setup describes a multi-round game with Player 2, suggesting a single partner that you interact with each round. If it was against a population, the description would likely refer to playing against a different or random opponent each round.}

\vspace{0.5cm}

\noindent\textbf{\underline{gpt-4.1}}:

Question response: {\fontfamily{qcr}\selectfont The description is framed as a two-player game, so you are playing against a single other player* If you were playing against a population, the payoff might be based on the frequency of choices in the population (e.g., evolutionary game theory), but here, your outcome depends only on the simultaneous choice of one other agent.}

\vspace{0.5cm}

\end{document}